\documentclass[onecolumn,12pt]{cohere}

\usepackage{lipsum}
\usepackage{stfloats} 
\usepackage{booktabs}
\usepackage{url}
\usepackage{hyperref}
\usepackage[hyphenbreaks]{breakurl}
\usepackage{graphicx}
\usepackage{multirow}
\usepackage{adjustbox}
\usepackage{float}
\usepackage{caption}
\usepackage{subcaption}
\usepackage[colorinlistoftodos]{todonotes}
\usepackage{lscape}
\usepackage{rotating}
\usepackage{enumitem}
\usepackage{wrapfig}
\usepackage{graphicx}
\usepackage{enumitem}
\usepackage{booktabs}
\usepackage{CJKutf8}
\usepackage{multirow}
\usepackage{amsfonts}
\usepackage{amsmath}
\usepackage{xcolor, colortbl}
\usepackage{multirow, hhline}
\usepackage{subcaption} 
\usepackage{arydshln}
\usepackage{siunitx}

\usepackage{nicematrix}

\title{The Reality of AI and Biorisk}

\author{
    name={Aidan Peppin},
    affiliation={Cohere For AI},
    email={aidanpeppin@cohere.com}
}
\author{
    name={Anka Reuel\footnotemark[1]},
    affiliation={Stanford University}
}
\author{
    name={Stephen Casper},
    affiliation={MIT}
}
\author{
    name={Elliot Jones},
    affiliation={Ada Lovelace Institute}
}
\author{
    name={Andrew Strait},
    affiliation={Ada Lovelace Institute}
}
\author{
    name={Usman Anwar},
    affiliation={University of Cambridge}
}
\author{
    name={Anurag Agrawal},
    affiliation={Ashoka University}
}
\author{
    name={Sayash Kapoor},
    affiliation={Princeton University}
}
\author{
    name={Sanmi Koyejo},
    affiliation={Stanford University},
}
\author{
    name={Marie Pellat},
    affiliation={Mistral AI},
}
\author{
    name={Rishi Bommasani\footnotemark[1]},
    affiliation={Stanford University},
}
\author{
    name={Nick Frosst},
    affiliation={Cohere and Cohere For AI}
}
\author{
    name={Sara Hooker},
    affiliation={Cohere For AI},
    email={sarahooker@cohere.com}
}
\abstract{To accurately and confidently answer the question "could an AI model or system increase biorisk”, it is necessary to have both a sound theoretical threat model for how AI models or systems could increase biorisk and a robust method for testing that threat model. This paper provides an analysis of existing available research surrounding two AI and biorisk threat models: 1) access to information and planning via large language models (LLMs), and 2) the use of AI-enabled biological tools (BTs) in synthesizing novel biological artifacts. We find that existing studies around AI-related biorisk are nascent, often speculative in nature, or limited in terms of their methodological maturity and transparency. The available literature suggests that current LLMs and BTs do not pose an immediate risk, and more work is needed to develop rigorous approaches to understanding how future models could increase biorisks. We end with recommendations about how empirical work can be expanded to more precisely target biorisk and ensure rigor and validity of findings.

}
\begin{document}
\footnotetext[1]{This work is unrelated to the involvement of this author in the EU AI Act Code of Practice.}
\newpage
\section{Introduction}

\begin{quote}
    \textit{To measure is to know.} \textbf{--- Lord Kelvin}
\end{quote}
A major focus of current efforts to govern the risks associated with AI models and technologies is concerned with biological risks, known as “biorisks”. These are risks that \textit{a biological event – such as [...] a release of a biological agent or biological material – adversely affects the health of humans, non-human animals, or the environment} \citep{lentzos_2022}. A relatively recent flurry of media reporting about the potential for AI to accelerate such biorisks, often featuring influential figures like think tank or industry CEOs, has influenced public discourse and evolved into policy and governance attention and activity \citep{metz_dozens_2024,reuters_us_2023, goode_national_2024, lovelace_ai-powered_2022}.

For example, the AI Safety Institutes in the US and UK are developing and conducting biorisk-related tests and guidance for advanced AI models \citep{uk_ai_safety_institute_ai_2024, nist_managing_2024}. Some AI model developers are evaluating their systems for biorisk, as well as partnering with external biology labs to test how AI models can be used safely in their work \citep{anthropic_frontier_2023, openai_building_2024, google_deepmind_alphafold_2024, openai_openai_2024}. Emerging legislative frameworks include specific provisions or references to biorisk. For example, the US White House Executive order specifies “biological threats” as a focus for testing of AI models, and includes a lower compute threshold for AI models trained using primarily biological sequence data \citep{house_executive_2023}. The EU AI Act notes biological risks in the context of “systemic risks” that may be associated with general-purpose AI models \citep{european_parliament_regulation_2024}. Considerations of biorisk also featured at the international AI Safety Summits at Bletchley Park, UK \citep{uk_government_bletchley_2023} and Seoul, South Korea \citep{uk_government_new_2024}. This recent focus has been motivated by the notion that some AI models or systems could \emph{amplify} biorisks.

\textbf{In this work, we review the available literature to-date to assess whether currently available evidence supports this focus and resource.}  Given the significant focus and resources allocated to the consideration of biorisk -- the claims and evidence underpinning it merit scientific scrutiny. Understanding whether the focus on biorisk is justified requires answering the question: \textit{could a specific AI model or system amplify biorisk?} To answer this question it is necessary to have both a sound theoretical threat model for how an AI model or system increases biorisk, as well as a robust method for testing the viability or likelihood of that theoretical threat model. Additionally, the theoretical threat models should clearly define and specify in detail the type of AI model(s) or system(s) of concern. Therefore, this paper asks: given publicly available evidence, \textbf{1)} are the current threat models sound and \textbf{2)} are the methods we have to test them robust?

\textbf{We find that studies around AI-related biorisk are nascent and often speculative in nature or limited in terms of their methodological maturity, and transparency around the methods underpinning some studies is limited.} This leads to uncertainty about the theoretical models for how AI may augment biorisk and empirical assessments of them, and therefore limits understanding of what appropriate, scientifically robust approaches to evaluating and mitigating biorisk may look like. Based upon the studies to-date and current capabilities of AI models, the popular concerns surrounding AI and biorisk are not  supported by the available scientific evidence. Furthermore, as AI capabilities increase, we recommend some important changes to increase the rigor and scientific grounding of approaches to understanding AI and biorisk:

\begin{enumerate}
    \item \textbf{Focus on whole-chain risk analysis}, considering how LLMs and BTs interact with the various complex stages of developing and deploying a harmful biological substance, including access to materials, specialized skills, laboratory facilities, etc, rather than solely focusing on assessments of AI models’ biological capabilities.

    \item \textbf{Direct attention towards AI models that are developed specifically for biological purposes,} such as biological tools and LLMs trained or fine-tuned to show high-performance on tasks specifically relevant to specific stages in the biorisk chain, rather than all general purpose AI models.

\item \textbf{Target policy and risk mitigation measures towards establishing more precise and accurate threat models} for how AI models uplift risk of physical harm, and developing robust empirical assessments for example with clear control variables and high-ecological validity, rather than relying on or mandating assessments that lack theoretical validity or methodological rigor. 
\end{enumerate}

\section{Background}
Biorisk research and governance has emerged progressively since the early 20th Century, with the 1925 Geneva Protocol responding to the use of biochemical weapons in the First World War, followed later by the 1972 Biological Weapons Convention \citep{united_nations_office_for_disarmament_affairs_history_noyear}. Beyond state development and use of bioweapons, national and international frameworks for biorisk management have iteratively emerged too, partly in response to terror group-led attacks, such as the Japanese doomsday cult Aum Shinrikyo \citep{kaplan_cult_noyear} or the Amerithrax attacks \citep{the_united_states_department_of_justice_amerithrax_2010}, as well as naturally emerging pandemics such as bird flu and Covid-19 \citep{cummings_emerging_2021}. These biorisk management frameworks largely center on implementing appropriate safety and security practices across facilities that conduct biological research and manufacturing, as well as monitoring of certain biological materials and synthetic development techniques \citep{national_science_and_technology_council_evidence-based_2022}. This is to reduce the likelihood of both accidental and malicious biological harms, by addressing risks across different stages in the “biorisk chain”. A typical “biorisk chain” involves multiple stages: an actor having an “irresponsible, misguided, or malicious” intention, which develops into a “biological idea", which is then "converted into biological data, such as a pathogen genome,” transforming the data into a “live biological artifact”, culturing and testing this artifact, before successfully dispersing it “into the target environment” \citep{sandberg_who_2020}.

Following recent advancements in AI technologies, particularly modern large language models based on transformer architecture \citep{vaswani_attention_2023} and biological models such as AlphaFold \citep{jumper_highly_2021}, attention has turned to how the novel capabilities offered by these technologies interact with the biorisk chain \citep{callaway_could_2024}. Publicly available research studies concerned with the novel capabilities of AI technologies have focused on their “dual use” nature in that they are “intended for benefit, but might easily be misapplied to do harm” \citep{world_health_organisation_what_2020}. These studies have explored theoretical threat models by which some types of AI model may augment biorisk. In the following section, we explore the available evidence around two of these threat models. 

\section{Assessing Amplification of Biorisk}

Most published work to-date on assessing the uplift in biorisk that comes from AI has centered around two threat models: \textbf{1) access to biological information and planning} and \textbf{2) synthesis of harmful biological artifacts}. For 1), amplification of risk is hypothesized to occur because Large language models (LLMs) could increase or “uplift” a users’ ability to gather information relevant to planning and carrying out biological attacks \citep{steph_batalis_ai_2023, rose_near-term_2024, gregory_c_allen_advanced_2023, rocco_casagrande_statement_2023}. For 2), the uplift in risk is hypothesized to be due to specialized AI "biological tools" assisting malicious actors in identifying new toxins, designing more potent pathogens, or optimizing existing biological agents for increased virulence \citep{nazish_jeffery_bio_2023, jeff_alstott_preparing_2023, halstead_managing_2024}.

For each of the threat models considered, we explore publicly available evidence around each of these threat models according to the following format:
\begin{enumerate}
    \item \textbf{Threat Model}: We introduce the assumed theoretical model for how a type of AI model may augment biorisk.
    \item \textbf{Relevant Research}: We present a summary of research, experiments, and/or studies conducted to-date around this proposed threat model. 
    \item \textbf{Assessment}: We evaluate the scientific conclusions we can draw about this threat model and what gaps or limitations in understanding remain, based on the available evidence.
\end{enumerate}

\subsection{Access to Biological Information and Planning}

\subsubsection{Threat Model}

\textbf{Hypothesized source of amplified threat:} Large language models (LLMs) could increase or “uplift” a users’ ability to gather information relevant to planning and carrying out biological attacks \citep{steph_batalis_ai_2023, rose_near-term_2024, gregory_c_allen_advanced_2023, rocco_casagrande_statement_2023}. 

This threat model draws from an established concern in bio-security literature known as the \textit{information access} threat model, where access to relevant knowledge provides an actor with an increased ability to conduct a biological attack \citep{cummings_emerging_2021}. Where an LLM is determined to support this increased access to relevant biological information, it is termed an “uplift” in the ability to carry out a biological attack \citep{openai_building_2024}. Here, the key question is whether access to LLMs fundamentally amplifies the degree of information access \emph{beyond what is already easily available} (e.g., through the internet). This increase in risk afforded by an AI model or system relative to risks afforded by existing available tools is often referred to as a "marginal risk” \citep{national_telecommunications_and_information_administration_dual-use_2024, kapoor2024societalimpactopenfoundation}.

\subsubsection{Relevant research} 

\begin{table}[t!]
\small
    \centering
    \begin{tabular}{>{\raggedright\arraybackslash}p{32mm}>{\raggedright\arraybackslash}p{15mm}>{\raggedright\arraybackslash}p{18mm}>{\raggedright\arraybackslash}p{21mm}>{\raggedright\arraybackslash}p{20mm}>{\raggedright\arraybackslash}p{19mm}>{\raggedright\arraybackslash}p{17mm}}
    \toprule
    \textbf{Name of Study} & \textbf{Sample Size}  &\textbf{Compared LLMs vs internet access?}&\textbf{Red Teamer Expertise}&\textbf{Findings} &\textbf{Models tested} &\textbf{Third Party Involvement}\\
    \midrule
    Can large language models democratize access to dual-use biotechnology?

\citep{soice_can_2023}&   9-12& No& Undergraduate& Potential for uplift& GPT-4, GPT 3.5, Bard, and FreedomGPT&None\\ \midrule
    The Operational Risks of AI in Large-Scale Biological Attacks: Results of a Red-Team Study

\citep{mouton_operational_2024}&   45& Yes& Varied levels of expertise with LLMs and biology& No statistically significant uplift& Not specified&Gryphon Scientific\\ \midrule
    An early warning system for LLM-aided biological threat creation. 

\citep{openai_building_2024}&   100& Yes& 50x biology PhDs, 50x undergraduate& No statistically significant uplift& GPT-4&Gryphon Scientific\\\midrule
 Claude 3 Model Card 

\citep{anthropic_claude_2024}& Not specified& Yes& Not Specified& Minor uplift (Unclear of statistical significance)& Claude 3&Gryphon Scientific\\
\bottomrule
    \end{tabular}
     \caption{\textit{Published information for existing red-teaming studies shows nascent methodological approaches. All studies which compare uplift to internet access find a non-significant increase in risk due to LLMs.}}
    \label{tab:study_methods}
\end{table}

Initial work on this topic by \cite{soice_can_2023} , evaluated how access to LLMs enables users to gather information about how to develop a pathogen or biological weapon, and plan to deploy it in the real world. Through a red teaming methodology, three groups of 3-4 students with no scientific training used LLMs to see how they could assist in planning and carrying out a biological attack, for example by using them to gather information about harmful biological artifacts, or to gain guidance on how to acquire those artifacts and deploy them to cause maximum harm. The authors said their findings “suggest that LLMs will make pandemic-class agents widely accessible [...] even to people with little or no laboratory training” \citep{soice_can_2023}. However, this study did not include a critical baseline – how access to information via LLMs compared to information access via, for example, sources on the internet.

In later work, this important baseline comparing LLMs and information available on the internet was added. Researchers at RAND Corporation applied a similar red teaming approach which involved 45 participants who had varying degrees of expertise with both LLM technologies and in biology. In contrast to the prior study, these groups were randomly assigned to have access to the internet and an LLM, or to the internet only. The team’s plans to develop a bio-attack were scored and it was found that the groups with access to LLMs in addition to the internet did \emph{not} score significantly higher than those without access to an LLM \citep{mouton_operational_2024}. No groups had access to only an LLM \textit{without} internet access and so it remains unclear how strong performance is with only access to an LLM. 

Since these initial studies, similar red teaming assessments have been carried out by AI model developers. \cite{anthropic_claude_2024} reported "minor uplift" in relation to their Claude 3 model, but  the statistical significance and methodological details are not fully reported. Work by \cite{openai_building_2024} included 100 red-teamers (more than 2x that of \cite{mouton_operational_2024}) with different levels of expertise to represent varied types of threat actor, and found no statistically significant uplift in threat actor capability. (See Table \ref{tab:study_methods}.)  important to note that the studies reviewed here highlight their own methodological limitations, for example the relatively limited sample sizes in and variety of expertise in terms of red team participants. Furthermore, we observe in Table \ref{tab:study_methods} that the majority of studies were conducted by the same third-party provider Gryphon
Scientific \footnote{Gryphon Scientific was acquired by Deloitte in April 2024. See: \url{https://www2.deloitte.com/us/en/pages/consulting/solutions/biotech-consulting-and-data-analytics-solutions.html}}. In particular, \cite{soice_can_2023} and \cite{mouton_operational_2024} position findings as exploratory, rather than conclusive, and do not rule out future, more capable LLMs elevating the ability to understand, synthesize, and communicate biological information.

\textbf{Access to biological knowledge is only one part of a biorisk chain.} Regarding future LLMs, even if they reduce barriers to entry for accessing information about how to create a harmful biological artifact, it remains an open question how this alters the overall likelihood of executing a successful bio-attack. Information access is typically only an early stage in the biorisk chain, and harms cannot manifest until malicious users synthesize a hazardous biological artifact and disperse it in the real world. This requires not only access to information but execution of a series of phases in the biorisk chain that LLMs may or may not augment \citep{nazish_jeffery_bio_2023, dewey_murdoch_written_2023, steph_batalis_ai_2023, nist_managing_2024}. While access to information is an important threat model to study, it is important to note that this threat model does not account for other steps in the chain and the necessary resources and knowledge needed to complete them. OpenAI’s 2024 study notes this too: \textit{"information access alone is insufficient to create a biological threat} adding that studies of information access alone do \textit{not test for success in the physical construction of the threats"} \citep{openai_building_2024}. Specialist training and access to well-resourced labs is critical for leveraging biological information effectively and there remain formidable barriers to entry for malicious actors, for example, access to the necessary physical equipment and materials, as well as deep understanding of wet lab protocols necessary to synthesize and release a harmful biological artifact \citep{jefferson_synthetic_2014, grushkin_seven_2013}.  Some estimates suggest such skills and materials access is limited to only 30,000 people globally \citep{righetti_towards_2024, esvelt_credible_2022}.

This means that even if access to information is uplifted by LLMs, the likelihood of a bio-attack may remain low due to these dependencies. To our knowledge, there have been no empirical studies which assess the connection between access to information via LLMs and other steps in the biorisk chain beyond information access. Additionally, recent work shows as yet unresolved challenges related to getting general foundation models to perform better than narrow, domain-specialized models \citep{jeong2024medicaladaptationlargelanguage, xu2024specializedfoundationmodelsstruggle}. This remains a significant gap in our understanding of how LLMs may augment biorisk through information and planning uplift, and the evidence which does exist suggests there is not statistically significant uplift.

\subsubsection{Assessment}

The available evidence suggests that information access via current, publicly available LLMs does not meaningfully increase the risk that an actor could plan and conduct a biological attack, compared to them simply having access to the internet. This is echoed by the United States’ National Security Commission on Emerging Biotechnology, which concluded in January 2024 that "At this time, LLMs do not significantly increase the risk of the creation of a bioweapon as LLMs do not provide new information [...] beyond what is already available on the internet" \citep{national_security_commission_on_emerging_biotechnology_white_2024}. Additionally, access to biological information is only one part of the risk chain, and harms cannot materialize until biological artifacts are physically tested and released. This means that the locus of risk lies not only in access to information via an LLM, but across the biorisk chain. Currently available evidence does not yet offer detailed theoretical models or empirical analyses for how biological information access and planning meaningfully increases risk across the whole chain.

To strengthen collective understanding of this threat model and appropriately target policy and risk mitigation measures, AI safety and governance researchers can work to develop more robust theoretical models for and studies of the \textit{link between} access to information via (increasingly capable) LLMs and the likelihood of real-world harm manifestation through the biorisk chain.

\subsection{Synthesis of harmful biological artifacts}

\subsubsection{Threat model}
\textbf{Hypothesized source of amplified threat:}  Specialized AI "biological tools" assist malicious actors in identifying new toxins, designing more potent pathogens, or optimizing existing biological agents for increased virulence \citep{nazish_jeffery_bio_2023, jeff_alstott_preparing_2023, halstead_managing_2024}.

This threat model is explicitly focused on the ability of malicious actors to use specialized AI models trained specifically on biological data and intended for application in biosciences to facilitate the creation of harmful biological artifacts. Therefore, in this section we ask: \textit{do AI models designed for use in bioscience make it easier to design harmful biological artifacts? }To do so, we first assess how specialized AI tools can be applied to biological sciences. Then we ask what empirical evidence there is that those tools could be applied to successfully augment the likelihood of a malicious actor carrying out a biological attack. 

\subsubsection{Relevant research}

Often termed together as “biological tools” (BTs) \citep{halstead_managing_2024, rose_near-term_2024}, a range of AI models and systems are being developed using biological data to perform tasks that support biological research and engineering. The majority of biological sequence models are trained on protein (amino acid), DNA or RNA sequences. This is often for tasks such as:
\begin{enumerate}
     \item \textbf{Protein structure prediction}, to model protein structures, functions, and interactions to improve understanding of cellular functions and aid the design of potential therapeutics \citep{abramson_accurate_2024, baek_efficient_2023}
    \item \textbf{Protein Design}, to model protein binders to aid protein engineering problems in relation to, for example, therapeutics, biosensors and enzymes \citep{dauparas_robust_2022}
    \item \textbf{Gene Element Prediction}, via foundation models that learn generalizable features from genome data, to predict how DNA changes affect an organism’s fitness and design biological systems \citep{nguyen_hyenadna_2023}
    \item \textbf{Pathogenic Variant Prediction}, by modeling of possible changes to amino acid chains to understand the affects for pathogenicity \citep{cheng_accurate_2023}
    \item \textbf{General-purpose biological sequence modeling}, which generates possible proteins in response to natural-language prompts, to aid in biological programming \citep{hayes_simulating_2024}.
\end{enumerate}

Core to understanding the dual-use nature of these models lies in determining how transferrable their biological use cases are to malicious settings. Applying AI models to biological problems is in its nascency: even in the case of AlphaFold \citep{jumper_highly_2021}, several studies have found that it performs less well than experimental structures as targets for the computational docking algorithms used in drug design \citep{read_alphafold_2023, terwilliger_alphafold_2023, karelina_how_2023}. The most serious limitations arise because these models "are based on learning patterns and know almost nothing about physics and chemistry" and "cannot consider factors such as pH, temperature or the binding of ions, other ligands or other proteins" \citep{read_alphafold_2023}.  Several papers have pointed out the difficulties of translating AlphaFold predictions outside of simulations, concluding that "despite the large recent gains in structure prediction accuracy, using predicted structures effectively for pharmaceutical applications remains a challenge" \citep{tourlet_alphafold2_2023, terwilliger_alphafold_2023}. The brittle and often clear shortcomings of current BTs illustrate that the danger as a dual-use tool is currently limited. 

\textbf{What does this mean for predicting unknown toxins or proteins that underpin harmful biological artifacts?} Designing novel diseases or bioweapons is considerably difficult \citep{european_commission_synthetic_2005, national_academies_press_biotechnology_200}. Skilled and technical expertise is required to utilize BTs effectively, such as cell culturing skills or experience working with genomes, in addition to hard-to-obtain or expensive material as well as organizational and staff resources \citep{national_academies_press_biotechnology_200, cornellBarriersBioweapons} (See Table \ref{tab:task_taxonomy}). This suggests that these risks are primarily limited to state actors or sophisticated users with advanced knowledge in biology and machine learning, reducing both scale of risk and the “attack surface” that needs to be monitored \citep{rose_near-term_2024, cornellBarriersBioweapons,dang2024ayaexpansecombiningresearch}. 

\begin{table}[t!]
\small
    \centering
    \begin{tabular}{>{\raggedright\arraybackslash}p{36mm}>{\raggedright\arraybackslash}p{41mm}>{\raggedright\arraybackslash}p{43mm}>{\raggedright\arraybackslash}p{22mm}}
 \textbf{Pathogen Type}& \multicolumn{2}{l}{\textbf{Requirements for Synthesis of Biological Artifact}}&\textbf{Barrier to Entry}\\
     & \textbf{Skills and Expertise}&\textbf{Material Resources}&\\
    \toprule
    Known Pathogenic Viruses & Common cell culture and virus purification skills.&Access to basic laboratory equipment, e.g. biosafety cabinet, cell culture incubator, centrifuge.&Medium \\ \midrule
    Known Pathogenic Bacteria & Specialized hands-on experience working with large bacterial genomes.&Significant financial and organizational resources.&High\\ \midrule
    Existing Viruses Modified & Advanced molecular biology skills and advanced knowledge of the field.&Basic to moderate resources similar to re-creating a known pathogenic virus.&Medium \\ \midrule
 Existing Bacteria Modified& Varied skill levels depending on bacterial modification, classical molecular biology expertise. &Basic resources similar to re-creating a known pathogenic virus.&Medium\\\midrule
 New Pathogens Created & Advanced design skills and tools. &Well-resourced teams with deep expertise in several different technologies; significant financial resources, and extensive testing capabilities.&High\\
    \end{tabular}
     \caption{Skill and resource barriers to synthesizing harmful biological artifacts are not overcome by the use of AI biological tools. (Adapted from \cite{NAP24890}).}
    \label{tab:task_taxonomy}
\end{table}

Even with state-sponsored expertise and resources, it is hard for experts to precisely target desirable characteristics in diseases such as contagiousness or stability, and when considering the necessary resources and skills to develop pathogens “the importance of tacit knowledge is commonly overlooked” \citep{lentzos_2022}. As researchers from the Soviet program to weaponize biological agents said: \textit{"}Everyone who has ever dealt with the genetics of bacteria knows how complicated it is to produce a new strain" \citep{leitenberg_soviet_2012}. In this sense, and similar to the information access threat model, discussed above, risks associated with BTs still only manifest at the point of synthesis of biological compounds in a real-world laboratory – not just simulation \textit{in silico} – and so further research is needed to clarify the theoretical model for how AI tools may reduce barriers to developing harmful biological artifacts with easy-to-access equipment or avoiding screening at laboratories and monitoring of biological materials \citep{nazish_jeffery_bio_2023, halstead_managing_2024, steph_batalis_ai_2023, terwilliger_alphafold_2023}.

Additionally, the efficacy of BTs is limited by data availability, meaning that BTs are limited in their potential for harm where they do not have access to data about harmful artifacts – or with intentional scrambling of such data \citep{campbell_censoring_2023, national_security_commission_on_emerging_biotechnology_white_2024}. Unlike many other problems within machine learning where datasets have grown rapidly in size and complexity over time \citep{longpre2023dataprovenanceinitiativelarge,singh2024ayadatasetopenaccesscollection,longpre2024consent}, BT data tends to be expensive to generate and fractured in access as many of the limited datasets are proprietary in nature \citep{the_royal_society_science_2024, goshisht_machine_2024, fuente-nunez_ai_2024}.

Given that access to data has dictated much of the progress in AI \citep{hooker_limitations_2024}, it is likely the rate of improvement of BTs will be far more fractured and harder to predict than other machine learning domains. This means even medium to longer term improvement in capabilities, as on the flip side risk is less likely to emerge at a predictable pace. 

\subsubsection{Assessment}

There are no known examples of current AI biological tools being misused to cause real-world harm, and complexities in the application of BTs to solve biological problems requires greater empirical research to understand how their misuse may manifest in real-world harms. This needed research includes whole-chain analyses that consider the risks associated with BTs in the context of biological artifact synthesis and deployment.

To strengthen collective understanding of this threat model, and appropriately target policy and risk mitigation measures, AI safety and governance researchers can expand empirical research and theoretical modeling for how BTs may interact across the entire lifecycle required to deploy a biorisk in the wild, rather than a sole focus on assessing BTs capabilities in the abstract. Studies must also report on important baselines such as how risk is amplified relative to access to existing tools, including the internet, and how a model trained only on data before a given cut-off would be able to perform on biology problems beyond that data. 

\section{Other potential threat models}
In addition to these two threat modes associated with AI and biorisk, some assessments have theorized that AI technologies may also augment biorisk when applied in code-generating tools which could be used to target autonomous labs (potentially powered by AI-based agentic systems) and (mis)direct their operations \citep{nazish_jeffery_bio_2023, halstead_managing_2024, inagaki_llms_2023}. To our knowledge, there is no publicly-available empirical or experimental research around this threat model. 

Other, analyses highlight that LLMs may considerably or significantly uplift malicious users’ ability to develop harmful biological artifacts through improving laboratory experimentation and troubleshooting, and reducing barriers to biological work, for example by enabling technical coding for computational biology to be performed with non-technical, natural language \citep{rose_near-term_2024}. There are few empirical studies of this threat model, though one example is reported in OpenAI’s o1 system card which assesses model capabilities across a range of biological tasks, such as troubleshooting wet lab protocols and automating wet lab work. From these evaluations, OpenAI reports that "o1 can help experts with the operational planning of reproducing a known biological threat." However, considering other barriers to threat creation, they conclude that o1-preview and o1-mini pose “limited risk,” because the models do not provide biology experts with new knowledge, and "do not enable non-experts to create biological threats, because creating such a threat requires hands-on laboratory skills that the models cannot replace” \citep{openai_openai_2024-1}. With regards to AI tools assisting lab-based work, it is notable that existing risk frameworks focus on mitigating physical synthesis of threats, for example focusing on containment and control procedures that secure the handling and production of biological artifacts, rather than digital capabilities that support knowledge gathering, task automation, or planning \citep{us_department_of_health_and_human_sciences_biosafety_2015}.

Relatedly, existing compute-based thresholds set, for example, in the White House Executive Order imply that the amount of compute used to train an AI model also contributes to the biorisks posed by AI models. However, establishing thresholds for what models constitute such higher-risk is a non-trivial task. There are several limitations of such thresholds, including that greater compute does not always yield increased capability, with smaller models meeting or exceeding larger models’ performance, and that the definition of “biological model” – often determined by the percentage of biological data a model is trained on – can be easily manipulated  \citep{hooker_limitations_2024, maug_biological_2024}. Additionally, the lower compute of developing BTs relative to larger, general purpose frontier models makes compute-based thresholds impractical to implement for biological tools \citep{moulange_towards_2023}.

\section{Conclusion: the way forward for AI and biorisk}

\begin{quote}
\textit{Extraordinary claims require extraordinary evidence -- Carl Sagan}
\end{quote} 

\textbf{Given evidence to-date, are current threat models for how AI models could augment biorisk sound, and are methods to test those threat models robust?} We have found that the available literature does not support the notion that access to biological information and planning via current, publicly available LLMs can significantly increase biorisk. While this does not offer conclusions for future models, more work is needed to develop this theoretical threat model before it can be considered viable or useful for assessing risks of AI models with additional capabilities. This includes accounting for how this threat model interacts with other stages throughout the entire the biorisk chain – for example enabling the acquisition of materials to physically synthesize a harmful biological artifact – as well as more clarity around the specific biological datasets or task-based training an LLM would need to demonstrate relevant capabilities. Current evidence suggests that LLMs lacking specialized biological data or capabilities and which cannot interact with other stages in the biorisk chain are unlikely to present risk.

For the second threat model considered in this paper, the synthesis of novel harmful artifacts through AI biological tools (BTs), we have found that current understanding of BTs’ capabilities and available evidence suggests that BTs still underperform in unexpected ways at core and known tasks, and so present limited risk in terms of their ability to propose novel formulations of harmful biological artifacts such as pathogens or toxins. Additionally, barriers to the availability of and access to data needed to train BTs suggest development of capabilities will be irregular, hard to predict, and slower than in other domains. Furthermore, empirical research is lacking to understand how this risk model may interact with the entire biorisk chain and manifest into real-world harm. 

Other potential threat models, such as the application of AI in code-generating tools for autonomous labs and the improvement of laboratory experimentation through LLMs, have little available theoretical or empirical research. It is notable that much of the available evidence is drawn largely from computer science fields: while some research draws on biorisk expertise, there is little cross-citation between research papers produced by the relatively nascent AI safety field and the existing research and practices of international biorisk management. In general, this suggests much more technical work is needed to provide empirical evidence and scientific support for some of the concerns around bio-risk \citep{reuel2024openproblemstechnicalai}.

\textbf{Given this assessment, we offer the following considerations for AI safety and governance researchers working across industry, academia, and government:}

\begin{enumerate}

\item Focus on whole-chain risk analysis, considering how LLMs and BTs interact with the various complex stages of developing and deploying a harmful biological substance, including access to materials, specialized skills, laboratory facilities, etc, rather than solely focusing on assessments of AI models’ biological capabilities.

\item Direct attention towards AI models that are developed specifically for biological purposes, such as biological tools and LLMs trained or fine-tuned to show high-performance on tasks specifically relevant to stages in the biorisk chain, rather than all general purpose AI models.

\item Target policy and risk mitigation measures towards establishing more precise and accurate threat models for how AI models uplift risk of physical harm, and developing robust empirical assessments for example with clear control variables and high-ecological validity, rather than relying on or mandating assessments that lack theoretical validity or methodological rigor. 
\end{enumerate}

Overall, the findings of this paper should not be misinterpreted as a dismissal of the potential for AI models and systems to augment biorisk – such dismissal would be irresponsible. However, our findings suggest that biological harms from AI models and systems remain a future risk rather than an immediate threat. It will be key to continually assess potential biorisks from new generations of advanced AI systems, particularly those trained on biological data. To better understand the level and nature of risk, these assessments must account for the above considerations. This will be crucial to supporting more precise and therefore effective and proportionate efforts by industry, government, and academia to assure AI safety in relation to biological applications.

\section{Acknowledgements}
We are grateful for valuable comments and feedback from: David Krueger, Markus Anderljung, Irene Solaiman, Yacine Jernite, Richard Moulange, and Miranda Bogen. 

\bibliography{main}

\end{document}